\documentclass[letterpaper, 10 pt, conference]{ieeeconf}  

\IEEEoverridecommandlockouts                              

\usepackage{graphics} 
\usepackage{epsfig} 
\usepackage{amsmath} 
\usepackage{amssymb}  
\usepackage[hidelinks]{hyperref}
\usepackage{subcaption}
\usepackage{multirow}
\usepackage{booktabs}

\title{\LARGE \bf
Booster Gym: An End-to-End Reinforcement Learning Framework for Humanoid Robot Locomotion
}

\author{Yushi Wang$^{1}$, Penghui Chen$^{1}$, Xinyu Han$^{2}$, Feng Wu$^{2}$, Mingguo Zhao$^{1}$
\thanks{$^{1}$Department of Automation, Tsinghua University, Beijing, China}%
\thanks{$^{2}$Booster Robotics Technology Co., Ltd, Beijing, China}%
}

\begin{document}

\maketitle
\thispagestyle{empty}
\pagestyle{empty}

\begin{abstract}
    Recent advancements in reinforcement learning (RL) have led to significant progress in humanoid robot locomotion, simplifying the design and training of motion policies in simulation. However, the numerous implementation details make transferring these policies to real-world robots a challenging task. To address this, we have developed a comprehensive code framework that covers the entire process from training to deployment, incorporating common RL training methods, domain randomization, reward function design, and solutions for handling parallel structures. This library is made available as a community resource, with detailed descriptions of its design and experimental results. We validate the framework on the Booster T1 robot, demonstrating that the trained policies seamlessly transfer to the physical platform, enabling capabilities such as omnidirectional walking, disturbance resistance, and terrain adaptability. We hope this work provides a convenient tool for the robotics community, accelerating the development of humanoid robots. The code can be found in \url{https://github.com/BoosterRobotics/booster_gym}.
\end{abstract}

\section{Introduction}

In recent years, reinforcement learning (RL) has emerged as a powerful technique for enabling humanoid robots to learn complex behaviors and tasks, especially in the area of locomotion. The ability to train motion policies in simulation has significantly reduced the complexity of designing robotic systems. However, translating these policies from a simulated environment to real-world robots remains a challenge due to a variety of factors, including the complexity of robot dynamics, sensory noise, and hardware limitations. To ensure successful deployment on physical robots, it is critical to design simulation environments and training procedures that promote the robustness and generalizability of the learned policies.

To address this challenge, we present a fully integrated code framework that encompasses the entire pipeline from simulation training to real-world deployment. The framework is compatible with customizable RL algorithms and reward function configurations, and incorporates domain randomization techniques to enhance policy robustness across varying conditions. Additionally, we address the challenge of handling parallel mechanical structures, which are common in humanoid robot design. We validate the proposed framework on the Booster T1 humanoid robot, demonstrating successful policy transfer from simulation to hardware. The trained policies enable the robot to execute tasks such as omnidirectional walking, disturbance resistance, and terrain adaptation. These results highlight the framework's practical utility in real-world applications. By releasing this framework as an open-source resource, we aim to provide a valuable tool for the robotics community to accelerate the development of humanoid robots.

\begin{figure}[!t]
    \centering
    \includegraphics[width=0.48\textwidth]{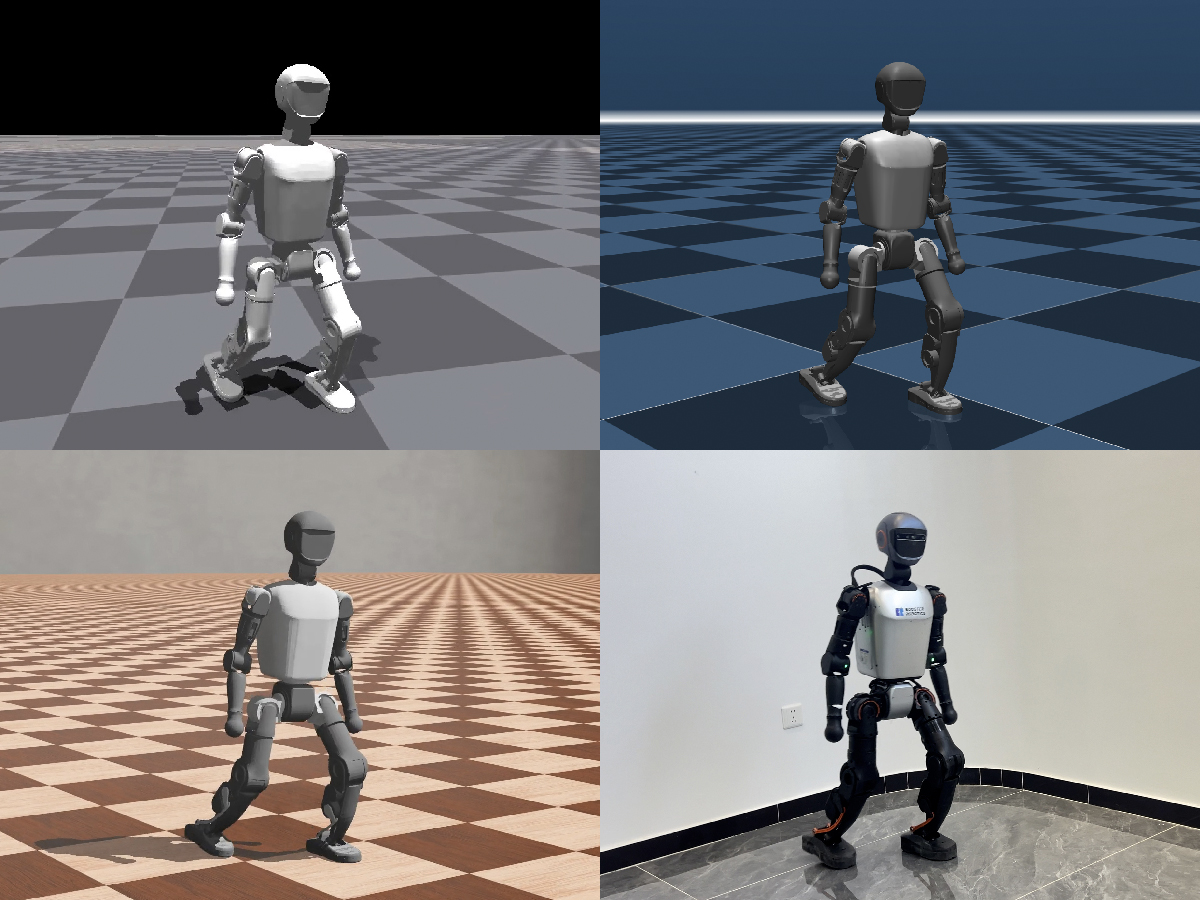}
    \caption{Training, testing, and deployment on Booster T1 across multiple environments. \textbf{Upper Left:} Training in Isaac Gym. \textbf{Upper Right:} Cross-Simulation testing in MuJoco. \textbf{Lower Left:} Verification in Webots via SDK. \textbf{Lower Right:} Deployment in the real world.}
    \label{fig:deploy}
\end{figure}

Our key contributions include:

\begin{itemize}
    \item A complete end-to-end solution for training and deploying RL-based locomotion policies, covering the entire pipeline from simulation to real-world deployment.
    \item Comprehensive domain randomization for environments, robots, and actuators, designed to reduce the sim-to-real gap and improve the robustness of trained policies when transferred to physical robots.
    \item Easily modifiable environment and algorithm interfaces, designed to allow researchers to efficiently adapt reward functions, network architectures, and physical parameters to meet diverse tasks.
\end{itemize}

\section{Related Works}

\subsection{Humanoid Locomotion Control}

With the rise of learning-based methods, the locomotion control for humanoid robots has experienced rapid development, achieving remarkable progress in multiple tasks. Traditional approaches, such as Whole-Body Control (WBC) \cite{li2023overview} and Model Predictive Control (MPC) \cite{khazoom2024tailoring}, rely on tracking handcrafted motion trajectories through model-based optimization. While these methods are effective in generating motions, they require extensive tuning of analytical models and often perform suboptimally in real-world settings due to external disturbances or inaccurate modeling. Additionally, real-time optimization on the robot makes these methods computationally intensive, necessitating workarounds like reduced precision or offloaded computations, which come with practical limitations.

In contrast, learning-based methods, particularly Reinforcement Learning (RL), have emerged as a powerful alternative for humanoid locomotion control, offering minimal modeling assumptions and enabling robots to acquire control policies through interactions with uncertain environments. This shift has facilitated the development of robust and adaptive locomotion controllers, capable of handling complex tasks such as standing up from diverse postures \cite{chen2025hifar} or navigating challenging terrains \cite{radosavovic2024real}.

Due to the high cost and risks of real-world training, policies are typically trained in simulated environments and deployed zero-shot on physical robots, making sim-to-real transfer a critical focus. Domain randomization has been widely adopted to bridge the sim-to-real gap by enhancing robustness through training policies across diverse simulated parameters such as dynamics, sensor noise, and environments \cite{tobin2017domain}. However, excessive randomization may lead to overly conservative policies and potentially hinder learning effectiveness \cite{chebotar2019closing}. To address this, leveraging historical information has emerged as a promising approach to infer task-specific strategies and improve policy generalization. For instance, methods like teacher-student distillation \cite{lee2020learning} and Rapid Motor Adaptation (RMA) \cite{kumar2021rma, kumar2022adapting} employ separated training stages. In these approaches, an expert policy is first trained with access to privileged information available only in simulation. This knowledge is then distilled into a deployable student policy, which relies solely on proprioceptive feedback and infers latent parameters from historical information, enabling real-time adaptation without explicit system identification. Beyond these multi-stage methods, recent advancements in end-to-end frameworks \cite{li2024reinforcement, gu2024advancing} achieve performance comparable to privileged expert policies while maintaining real-world deployability. Additionally, some studies integrate real-world data to refine modeling accuracy and reduce reliance on excessive randomization, balancing simulation efficiency with real-world applicability \cite{hwangbo2019learning}.

\subsection{Tools for Humanoid Learning and Simulation}

Deep reinforcement learning for robot is inherently computationally intensive, as it relies on extensive interactions between agents and their environments to explore effective policies. To address this challenge, large-scale parallelization has become a critical approach for reducing training time. While traditional robotic simulators like MuJoCo and Webots offer efficient and precise rigid body dynamics implementations, their reliance on CPU processing limits their ability to achieve massive parallelism. The introduction of GPU-accelerated physics simulators, such as NVIDIA Isaac Gym \cite{makoviychuk2021isaac}, has transformed this landscape by enabling physics simulation, reward and observation calculations, and neural network training to be executed entirely on the GPU. This innovation significantly accelerates training by supporting thousands of parallel environments and minimizing data-copying overhead \cite{rudin2022learning}. Building on the success of Isaac Gym, NVIDIA introduced Isaac Lab, which extends its capabilities into a more general-purpose robotics simulation platform with support for various tools. Isaac Lab demonstrates significant potential for advancing robot simulation, though it remains resource-intensive and currently has a limited user base. Recently, several other GPU-accelerated simulators, such as MuJoCo Playground \cite{zakka2025mujoco} and Genesis \cite{Genesis}, have also emerged, further driving progress in the field and expanding the range of available high-performance simulation tools.

In this work, we select Isaac Gym as the training environment due to its minimal yet sufficient implementation for humanoid locomotion tasks and is widely adopted in the robotics community. Although the PhysX physics engine used by Isaac Gym has certain limitations, such as the lack of support for closed kinematic chains and less accurate contact force estimation, we address these challenges through careful design choices to mitigate their impact. Additionally, we perform cross-simulation testing in MuJoCo and Webots to verify the generalization of the trained policies across different environments, following a similar approach to \cite{gu2024humanoid}. This multi-simulator strategy allows us to leverage the strengths of each platform, balancing high-efficiency training with lightweight and precise evaluation.

\section{Method}

\subsection{Reinforcement Learning Formulation}

We formulate the robot control problem as a Partially Observable Markov Decision Process (POMDP) and solve it via reinforcement learning (RL). The POMDP is defined as $\mathcal{M}=\langle\mathcal{S},\mathcal{A},p,r,\mathcal{O},\gamma\rangle$, where $\mathcal{S}$ and $\mathcal{A}$ are the state and action spaces, $p(\boldsymbol{s}'|\boldsymbol{s},\boldsymbol{a})$ is the state transition function, $r(\boldsymbol{s},\boldsymbol{a})$ is the reward function, $\mathcal{O}$ is the observation space, and $\gamma\in[0,1]$ is the discount factor. In the POMDP, the agent only has access to partial information about the complete state, which is consistent with information that can be obtained from sensors on the robot in the real world.

The agent aims to learn a policy $\pi(\boldsymbol{a}_t|\boldsymbol{o}_t)$  mapping observations to action distributions to maximize expectation of the discounted sum of future rewards:
\begin{equation}
J(\pi)=\mathbb{E}_{\tau \sim p_\pi} \left [ \sum_{t=0}^\infty \gamma^t r(\boldsymbol{s}_t, \boldsymbol{a}_t) \right ],
\end{equation}
where $\tau=(\boldsymbol{s}_0,\boldsymbol{a}_0,\boldsymbol{s}_1,\cdots)$ represents a trajectory of the agent sampled from the POMDP $\mathcal{M}$ under the policy $\pi$. 

We adopt an asymmetric actor-critic (AAC) \cite{pinto2017asymmetric} architecture with inconsistent observation spaces of the actor and critic, and apply Proximal Policy Optimization (PPO) \cite{schulman2017proximal} as the RL algorithm to trian the policy. At each iteration, the agent collects trajectories by executing the policy, then performs multiple steps of optimization using the same trajectory. The policy gradient is expressed as:
\begin{equation}
    \nabla_\theta J(\pi_\theta) = \mathbb{E}_{\tau \sim p_{\pi_{\theta_{\text{old}}}}} \left[\frac{\pi_{\theta}(\boldsymbol{a}_t|\boldsymbol{o}_t)}{\pi_{\theta_{\text{old}}}(\boldsymbol{a}_t|\boldsymbol{o}_t)}A^{\pi_{\theta_{\text{old}}}}  \nabla_\theta\log\pi_\theta(\boldsymbol{a}_t|\boldsymbol{o}_t)\right],
    \label{eq:policy_grad}
\end{equation}
where $\frac{\pi_{\theta}(\boldsymbol{a}_t|\boldsymbol{o}_t)}{\pi_{\theta_{\text{old}}}(\boldsymbol{a}_t|\boldsymbol{o}_t)}$ is the importance sampling weight to address the discrepancy between the current policy $\pi_{\theta}$ and the policy $\pi_{\theta_{\text{old}}}$ used to collect the data, and $A^{\pi_{\theta_k}}$ is an estimator of the advantage function under the policy $\pi_{\theta_k}$ calculated by Generalized Advantage Estimation (GAE) \cite{schulman2015high}.

The basic loss function of PPO is defined as:
\begin{equation}
    \mathcal{L}(\theta)=\mathcal{L}_{\text{policy}}(\theta)+c_{\text{value}}\mathcal{L}_{\text{value}}(\theta)-c_{\text{entropy}}H(\pi_\theta),
\end{equation}
where $\mathcal{L}_{\text{policy}}(\theta)$ is the policy surrogate loss, $\mathcal{L}_{\text{value}}(\theta)=\|V_\theta(\boldsymbol{s}_t)-\hat{V}\|^2$ is the value function error, $H(\pi_\theta)$ is the entropy regularization term to improve exploration, and $c_{\text{value}},c_{\text{entropy}}$ are coefficients. The surrogate loss $\mathcal{L}_{\text{policy}}(\theta)$ is clipped to constrain the magnitude of policy updates during training: 
\begin{equation}
    \begin{aligned}
    \mathcal{L}_{\text{policy}}(\theta)&=\mathbb{E}\left[\min\left(\frac{\pi_{\theta}(\boldsymbol{a}_t|\boldsymbol{o}_t)}{\pi_{\theta_{\text{old}}}(\boldsymbol{a}_t|\boldsymbol{o}_t)}A^{\pi_{\theta_{\text{old}}}},\right.\right.\\
    &\left.\left.\text{clip}\left(\frac{\pi_{\theta}(\boldsymbol{a}_t|\boldsymbol{o}_t)}{\pi_{\theta_{\text{old}}}(\boldsymbol{a}_t|\boldsymbol{o}_t)},1-\varepsilon,1+\varepsilon\right)A^{\pi_{\theta_{\text{old}}}}\right)\right].
    \end{aligned}
\end{equation}

\subsection{Problem Setup}
In the asymmetric actor-critic architecture, the actor and critic are represented by a policy network and a value network parameterized by $\theta$. The policy $\pi_\theta(\boldsymbol{a}_t|\boldsymbol{o}_t)$ determines the action $\boldsymbol{a}_t$ based on the partial observation $\boldsymbol{o}_t$. The value network is trained to estimate the state value $V_\theta(\boldsymbol{s}_t)$, receiving the full state $\boldsymbol{s}_t$, which is only available within the simulator.

\begin{figure}[!t]
    \centering
    \includegraphics[width=0.48\textwidth]{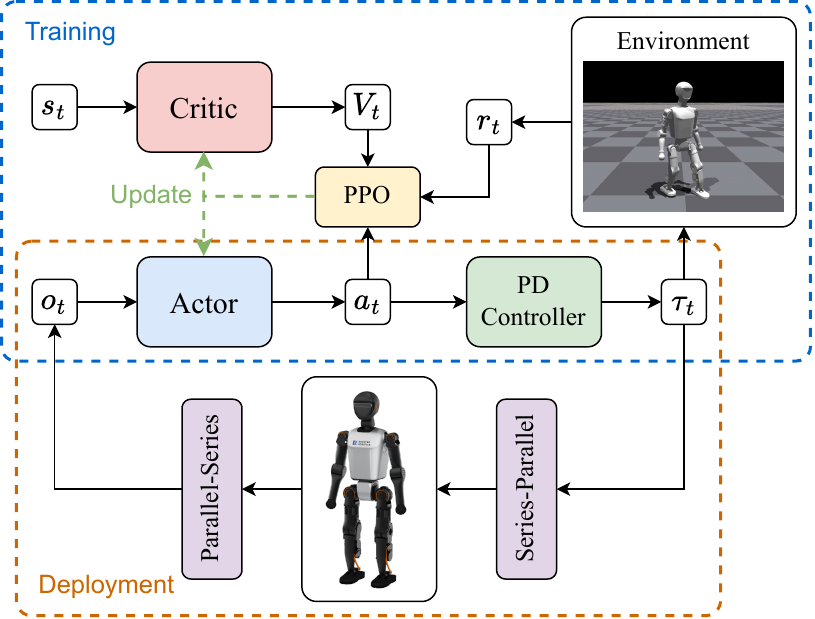}
    \caption{An overview of the control architecture for training and deployment. The actor and critic networks are optimized using PPO in a simulated environment during training. The actor network generates actions, which are subsequently converted to control signals via a PD controller.}
    \label{fig:frame}
\end{figure}

\subsubsection{Observation Space}
The proprioceptive observation of the robot includes the base angular velocity $\boldsymbol{\omega}_t$ and the gravity vector in the base frame $\boldsymbol{g}_t$ measured by an IMU, along with the joint positions $\boldsymbol{q}_t$ and velocities $\dot{\boldsymbol{q}}_t$ obtained from motor feedback. We construct the actor observation $\boldsymbol{o}_t$ by proprioceptive observation with noise, as well as the previous action $\boldsymbol{a}_{t-1}$ and the velocity command $(v_x,v_y,\omega_{\text{yaw}})$ for the omnidirectional walking task. Additionally, a gait cycle $(\cos(2\pi ft), \sin(2\pi ft))$ is included in the observation to guide the policy in learning a periodic walking gait \cite{siekmann2021sim}, where $f$ is the gait frequency. For the standing gait, the gait cycle is set to zero, enabling the policy to transition seamlessly between standing and walking. The critic input includes actor observation without noise and other privileged information to improve the value estimation. The details of observation are concluded in Table \ref{tab:observation}.

\begin{table}[!t]
    \centering
    \caption{Summary of Observation Space.}
    \label{tab:observation}
    \begin{tabular}{lccc}
    \toprule
    \textbf{Components} &\textbf{Dims} & \textbf{Actor} & \textbf{Critic} \\
    \midrule
    Commands & 3 & \checkmark & \checkmark \\
    Gait Cycle & 2 & \checkmark & \checkmark \\
    Gravity Vector & 3 & \checkmark & \checkmark \\
    Angular Velocity & 3 & \checkmark & \checkmark \\
    Joint Position & 12 & \checkmark & \checkmark \\
    Joint Velocity & 12 & \checkmark & \checkmark \\
    Previous Action & 12 & \checkmark & \checkmark \\
    \midrule
    Body Mass & 1 & & \checkmark \\
    Body Center of Mass & 3 & & \checkmark \\
    Base Linear Velocity & 3 & & \checkmark \\
    Base Height & 1 & & \checkmark \\
    Push Force & 2 & & \checkmark \\
    Push Torque & 3 & & \checkmark \\
    \bottomrule
    \end{tabular}
\end{table}

\subsubsection{Action Space}
The policy outputs joint position offsets as the action vector $\boldsymbol{a}_t$. The desired joint positions for the robot are determined by:
\begin{equation}
    \boldsymbol{q}_{\text{des}}=\boldsymbol{q}_{0}+\boldsymbol{a}_t,
\end{equation}
where $\boldsymbol{q}_{0}$ is default joint positions. Then, the desired joint positions are converted into torque commands directly by the PD controller on the motor driver with fixed gain and zero desired joint velocities:
\begin{equation}
    \boldsymbol{\tau}_{\text{des}}=\boldsymbol{k}_p(\boldsymbol{q}_{\text{des}}-\boldsymbol{q})-\boldsymbol{k}_d\dot{\boldsymbol{q}}
\end{equation}
The policy operates at a rate of 50 Hz, while the internal PD controller runs at a higher frequency, improving stability of the policy on the real robot compared to making the policy directly output torque.

\subsubsection{Reward Function}
Designing proper reward functions is important for obtaining desired behaviors for robots. We define the reward function as the weighted summation of several reward components:
\begin{equation}
    r(\boldsymbol{s}_t,\boldsymbol{a}_t)=\sum w_ir_i
\end{equation}
The reward function consists of three parts: tracking rewards, gait rewards, and regularization rewards.

To enable the robot to move in a specified manner, we define a torso velocity tracking reward. Additionally, we implement curriculum learning to progressively increase the magnitude of velocity commands, allowing the robot to learn walking more quickly. Commands for each environment are resampled within a random time window, and with a certain probability, the command is set to “stand still” to encourage the robot to learn transitions between standing and walking, as well as adjustments to different walking speeds. During training, when the command changes, we truncate the sequence to prevent the robot from making unrealistic predictions that could affect tracking performance.

For humanoid robot locomotion tasks, a gait reward is designed to encourage leg movement. We define a reference gait cycle and reward the robot for stepping with both legs according to the set cycle. Due to the simplified collision estimation in Isaac Gym, we use the difference in height between the foot and the ground, rather than foot contact forces, to determine whether the robot is lifting its leg.

Additionally, we define penalties for torso posture, energy consumption, and regularization terms to improve overall performance. For a detailed breakdown of the reward function settings, please refer to the table.

\begin{table}[!t]
    \centering
    \caption{Summary of Reward Function.}
    \begin{tabular}{lcc}
    \toprule
    \textbf{Components} &\textbf{Equations} & \textbf{Weights} \\
    \midrule
    Survival & 1 & 0.025 \\
    Velocity tracking (x) & $\exp(-(v_x^{\text{cmd}}-v_x)^2/\sigma_x)$ & 1.0 \\
    Velocity tracking (y) & $\exp(-(v_y^{\text{cmd}}-v_y)^2/\sigma_y)$ & 1.0 \\
    Velocity tracking (yaw) &$\exp(-(\omega_z^{\text{cmd}}-\omega_z)^2/\sigma_x)$ & 0.5 \\
    Base height & $(h^{\text{des}}-h)^2$ & $-20.0$ \\
    Orientation & $\|\boldsymbol{g}\|^2$ & $-5.0$ \\
    \midrule
    Torque & $\|\boldsymbol{\tau}\|^2$ & $-2\times 10^{-4}$ \\
    Torque tiredness & $\|\boldsymbol{\tau}/\boldsymbol{\tau}_{\max}\|^2$ & $-1\times 10^{-2}$ \\
    Power & $\max(\boldsymbol{\tau}\cdot\dot{\boldsymbol{q}},0)$ & $-2\times 10^{-4}$\\
    Lin velocity (z) & $v_z^2$ & $-2.0$ \\
    Ang velocity (xy) & $\|\boldsymbol{\omega}_{xy}\|^2$ & $-0.2$ \\
    Joint velocity & $\|\dot{\boldsymbol{q}}\|^2$ & $-1\times 10^{-4}$ \\
    Joint acceleration & $\|\ddot{\boldsymbol{q}}\|^2$ & $-1\times 10^{-7}$ \\
    Base acceleration & $\|\dot{\boldsymbol{v}}\|^2+\|\dot{\boldsymbol{\omega}}\|^2$ & $-1\times 10^{-4}$ \\
    Action rate & $\|\boldsymbol{a}_t-\boldsymbol{a}_{t-1}\|^2$ & $-1.0$ \\
    Joint position limit & $1_{\boldsymbol{q}>\boldsymbol{q}_{\max}}+1_{\boldsymbol{q}<\boldsymbol{q}_{\min}}$ & $-1.0$ \\
    Collision & $n_{\text{collision}}$ & $-1.0$ \\
    \midrule
    Feet swing & $1_{\text{feet swing}}\cdot1_{\text{swing period}}$ & $3.0$ \\
    Feet slip & $1_{\text{feet stance}}\cdot \|\boldsymbol{v}_{\text{feet}}\|^2$ & $-0.1$ \\
    Feet yaw & $\|\boldsymbol{\psi}_{\text{feet}}-\psi_{\text{base}}\|^2$ & $-1.0$  \\
    Feet roll & $\|\boldsymbol{\phi}_{\text{feet}}\|^2$ & $-0.1$  \\
    Feet distance & $\max(d_{\text{ref}}-d_{\text{feet}},0)$ & $-1.0$  \\
    \bottomrule
    \end{tabular}
\end{table}

\subsubsection{Episode Design}

The maximum episode duration is set to 1500 steps, equivalent to 30 s. The environment is reset when time limit is exceeded, but the subsequent rewards will be filled in the GAE. To prevent the agent from wasting time exploring ineffective states, such as when the robot has fallen, we implement additional early termination conditions, including excessively low base height or base velocity exceeds a threshold. Under these conditions, the environment is reset without granting subsequent rewards, thereby preventing the robot from falling or exhibiting extreme behaviors. Furthermore, the total reward for each frame is clipped to zero to avoid incentivizing early termination with negative rewards.

\subsection{Bridging the Sim-to-Real Gap}

Due to the limitations of real-world data collection in terms of efficiency and safety, reinforcement learning (RL) typically trains policies in simulation and deploys them in the real world with zero-shot transfer. However, the sim-to-real gap arises from the idealized assumptions in simulation, which fail to account for real-world uncertainties such as sensor noise, actuator dynamics, and environmental variations.

To address the sim-to-real gap, we leverage domain randomization to bridge the disparity between simulation and real-world deployment. For humanoid locomotion, we categorize these uncertainties into three primary sources: robot body dynamics, actuator characteristics, and environmental conditions. By applying domain randomization to each of these categories during simulation training, we enable the policy to adapt to a wide range of real-world scenarios.

Specifically, we randomize the mass and center of mass (CoM) positions of the trunk and other links, while introducing noise in observations to account for variations in the robot's physical properties. To mimic real-world actuator behaviors, we introduce randomization in joint stiffness, damping, and friction, as well as communication delays. Additionally, we simulate diverse terrains and randomize contact properties such as friction, compliance, and restitution to reflect environmental variations. To further enhance robustness, we randomly apply external disturbances, including kicks and pushes, by simulating sudden changes in velocity or applying forces over time. These disturbances help the policy adapt to unexpected real-world challenges, such as recovering from external impacts. Through these comprehensive strategies, our framework achieves robust sim-to-real transfer, allowing humanoid robots to perform complex locomotion tasks in diverse real-world environments with adaptation to different robot configurations.

\subsection{Deployment}

To facilitate the deployment of trained policies onto physical robots, we develop a Python-based deployment framework, accompanied by a pre-trained policy, allowing users to seamlessly implement and evaluate reinforcement learning (RL) policies on real hardware.

The trained policies are exported in a Just-In-Time (JIT) compiled format, ensuring efficient execution on the robot's onboard CPU. The JIT-optimized policy operates at a frequency of 50 Hz, processing real-time sensor data as input and generating joint position targets as output, which are then sent to the motors. The motor's PD controller is employed with predefined $K_p$ and $K_d$ gains, while joint velocity targets and feedforward torque are set to zero, aligning with the simulation configuration. This approach leverages the motor's built-in PD controller to achieve higher control frequency, rather than implementing a custom solution.

Additionally, we introduce the Booster Robotics SDK, a developer-focused toolkit that abstracts hardware interfaces, allowing external programs to interact with the robot through a unified API. The SDK is built on the Data Distribution Service (DDS) middleware, providing a Publisher-Subscriber communication model for efficient data exchange. It offers low-level motor control interfaces and access to sensor data, such as IMU measurements and joint encoder readings, enabling precise and direct interaction of the robot's hardware.

Parallel structures are a common design choice for humanoid robot ankles, as they help reduce the inertia of the foot and enhance dynamic performance. In our framework, we address the challenge of handling closed kinematic chains in parallel ankle mechanisms by implementing a series-parallel conversion module in the SDK. This module allows users to seamlessly train policies using a virtual serial structure, which is more computationally efficient and widely supported by current GPU-based physics engines, while enabling deployment on parallel mechanisms without additional complexity. Specifically, the module calculates position and velocity feedback for the virtual series joints using the robot's kinematic model and dynamically converts the policy outputs from the series model to the parallel structure using a transposed Jacobian matrix and a PD controller. This approach reduces the sim-to-real gap and simplifies deployment, ensuring smooth and accurate control.

\section{Experiment}

\begin{figure}[!t]
    \centering
    \includegraphics[width=0.48\textwidth]{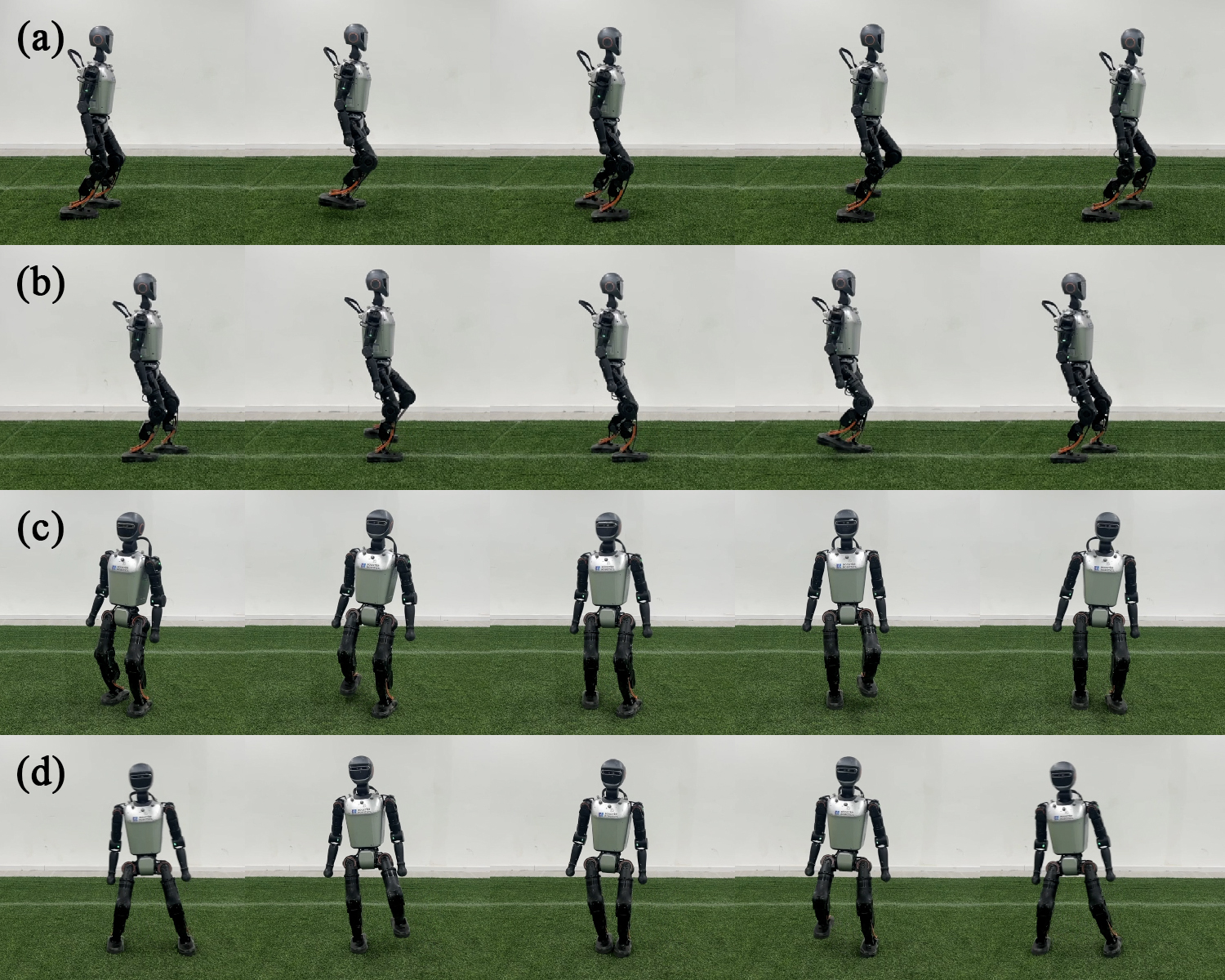}
    \caption{Omnidirectional walking. (a) Forward. (b) Backward. (c) Rotation. (d) Sideways.}
    \label{fig:omniwalk}
\end{figure}

\begin{figure}[!t]
    \centering
    \includegraphics[width=0.48\textwidth]{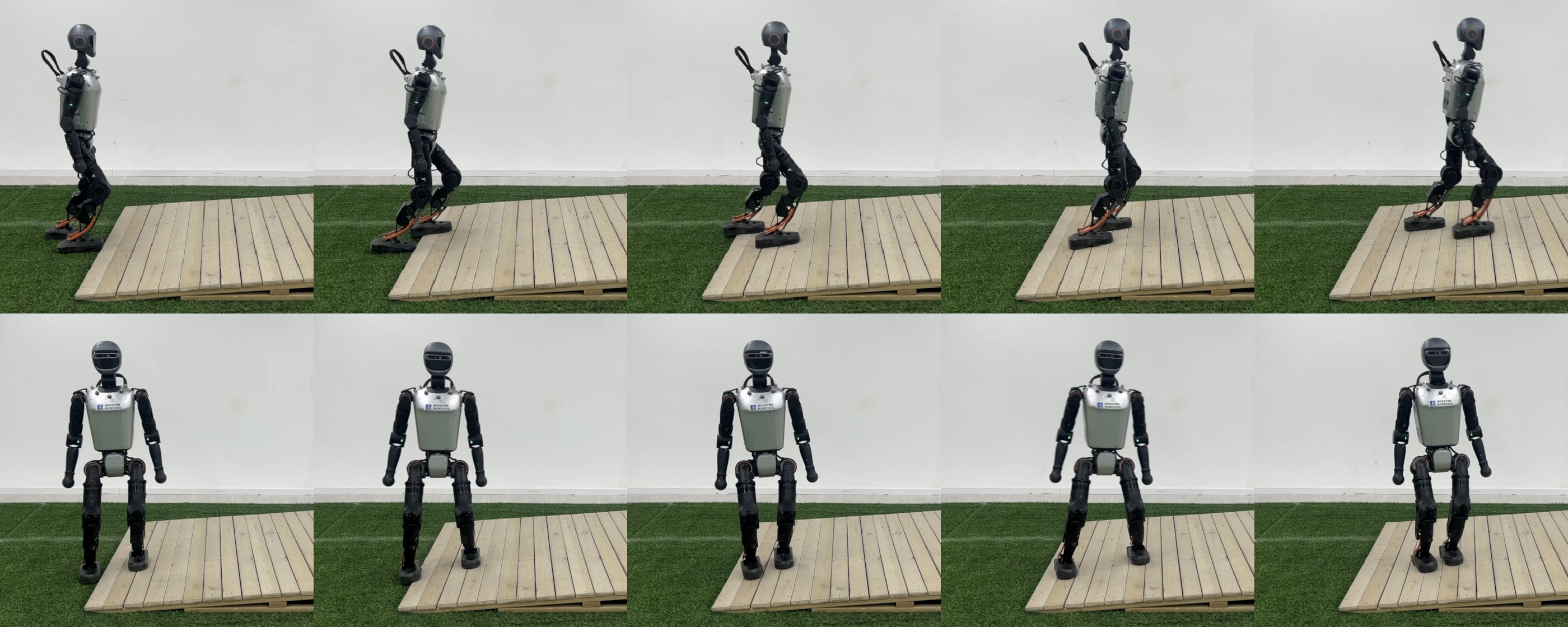}
    \caption{Walking on a 10-degree slope. \textbf{Upper:} Forward. \textbf{Lower:} Sideways.}
    \label{fig:slope}
\end{figure}

In this section, we evaluate the performance of policies trained using our proposed framework across a range of humanoid locomotion tasks by deploying them on the Booster T1 robot in real-world scenarios. The experiments are designed to assess key capabilities, including omnidirectional walking, adaptability to diverse terrains, robustness to external disturbances, and the effectiveness of sim-to-real transfer.

\subsection{Omnidirectional Walking Performance}

In this work, we focus on omnidirectional walking, the fundamental locomotion capability of humanoid robots, using a basic RL framework. We train the robot to follow linear velocity commands in the $x$ and $y$ directions and angular velocity commands in the yaw direction, issued via a joystick. The robot is rewarded for accurately tracking these velocity commands, promoting the development of omnidirectional movement.

As shown in Fig. \ref{fig:omniwalk}, we demonstrate the robot's ability to perform omnidirectional walking, including forward, backward, sideways, and rotational movements. The robot effectively tracks both single-direction and mixed-direction commands, as well as commands that change over time. These results highlight the effectiveness of our framework in handling fundamental locomotion tasks.

\subsection{Walking on Diverse Terrains and Under Disturbances}

\begin{figure}[!t]
    \centering
    \includegraphics[width=0.48\textwidth]{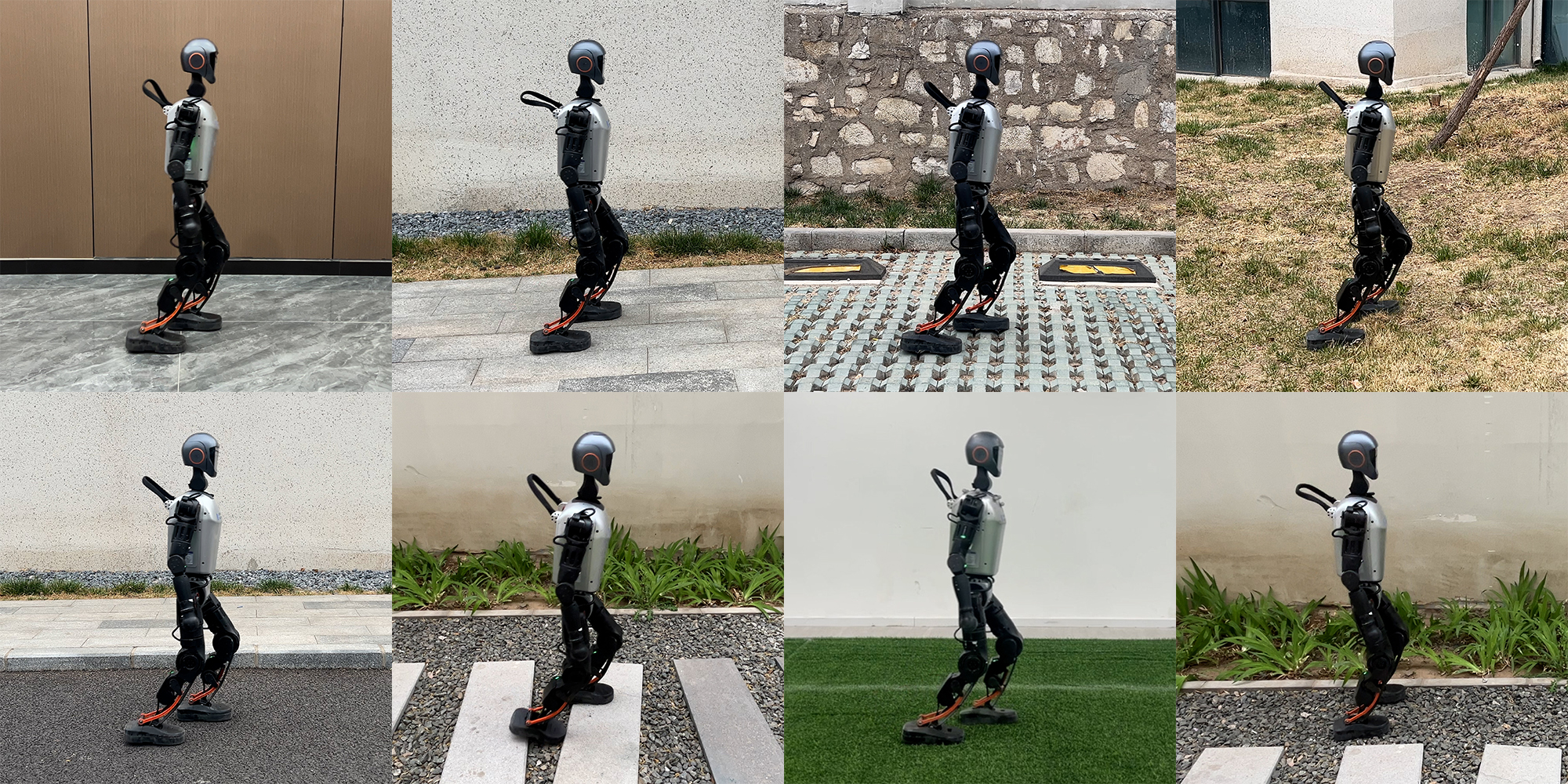}
    \caption{Walking on different types of terrains with the same policy.}
    \label{fig:terrain}
\end{figure}

\begin{figure}[!t]
    \centering
    \includegraphics[width=0.48\textwidth]{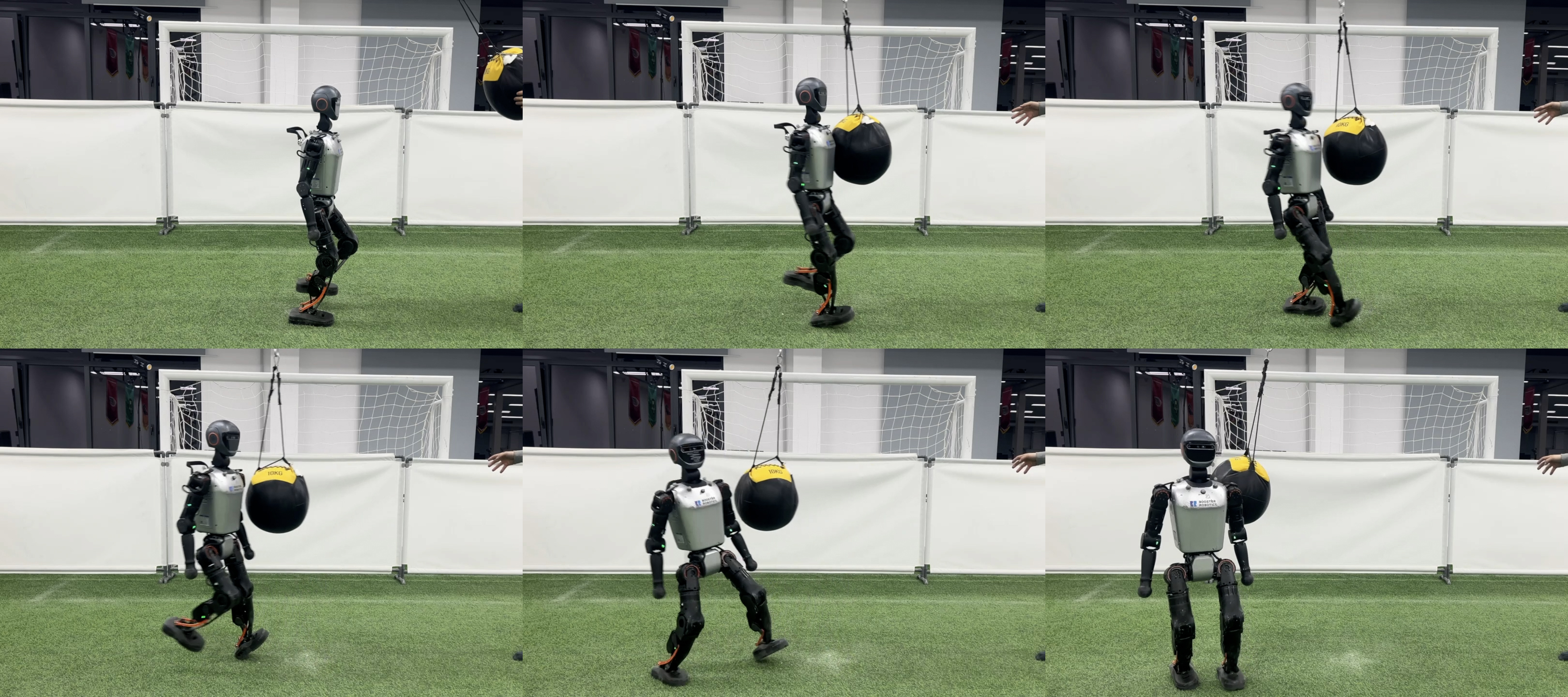}
    \caption{Push Recovery. A 10-kg ball impacts the robot during stepping in place. The robot regains a stable gait within a few steps after the disturbance.}
    \label{fig:push}
\end{figure}

\begin{figure*}[!t]
    \centering
    \includegraphics[width=\textwidth]{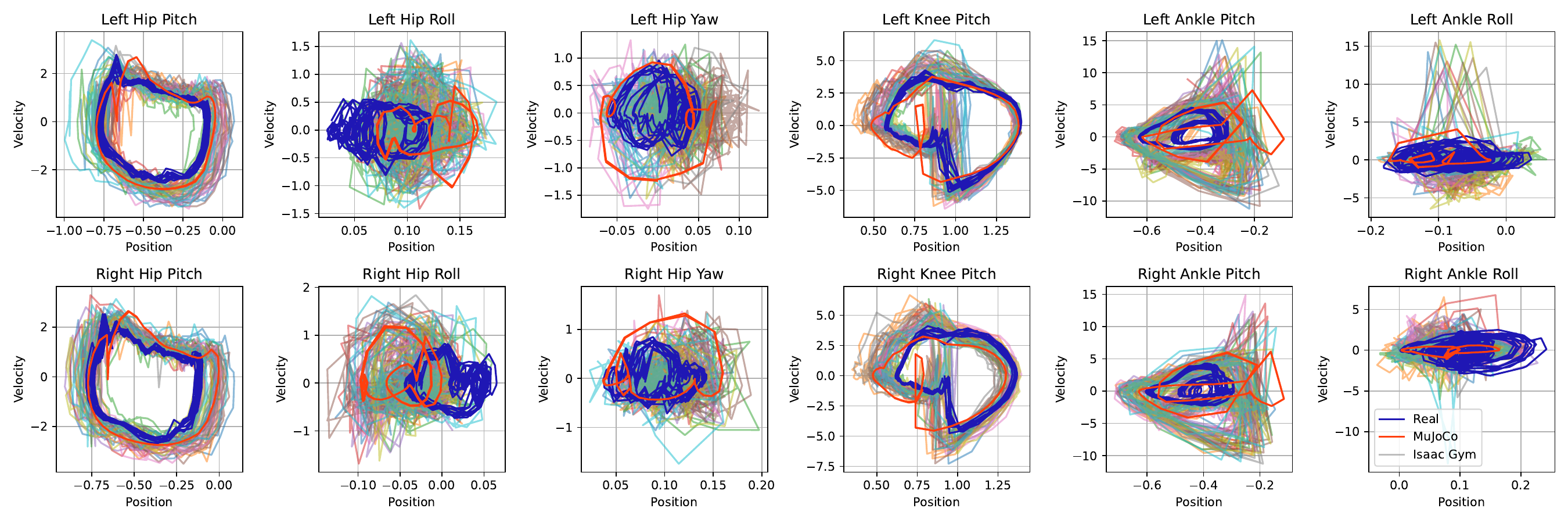}
    \caption{Joint position-velocity trajectories with a 0.5 m/s forward velocity commands in Isaac Gym, MuJoCo, and real-world deployment. The Isaac Gym trajectories represent results from 10 distinct environments sampled under domain randomization.}
    \label{fig:sim2real}
\end{figure*}

In this experiment, we evaluate the robustness of our policy by testing its performance across a variety of terrains and in the presence of external disturbances. As illustrated in Fig. \ref{fig:slope}, the trained policy enables stable locomotion on multiple surfaces, including grass, stone pathways, soil, asphalt, concrete, tiled floors. Notably, the policy demonstrates strong adaptability, even on challenging terrains. For instance, we construct a 10-degree slope, and the robot successfully performs forward, backward, and turning motions while adapting to step transitions, despite lacking explicit terrain perception.

We also assess the policy's resilience to external disturbances, simulating unexpected collisions the robot might encounter in real-world environments. This is tested through two scenarios: impulsive impacts and sustained forces. We introduce an instantaneous impact by dropping a 10 kg weight from a distance of 1 m while the robot is stepping in place. As shown in Fig. \ref{fig:push}, the impact causes significant displacement of the robot's torso. However, the policy swiftly responds by adjusting the foot placement, allowing the robot to regain stability within a few steps. Additionally, we test the robot's ability to withstand continuous external forces during forward motion. When an external force impedes its movement, the robot maintains stability and promptly resumes movement upon force removal.

While employing basic RL methods, our framework achieves notable real-world adaptability through carefully designed environmental variations in training. These results further highlight the robustness of the policy in handling real-world challenges.

\subsection{Sim-to-Real Transfer}

\begin{figure}[!t]
    \centering
    \includegraphics[width=0.48\textwidth]{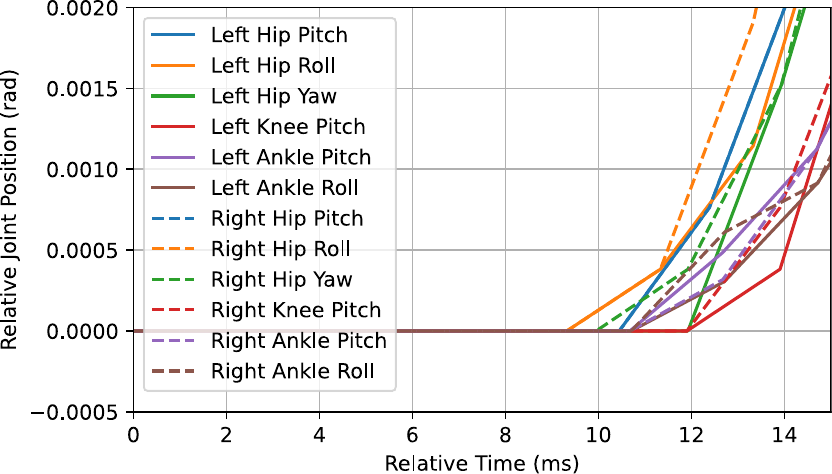}
    \caption{Joint position response to a 0.1 rad step command during the latency test. The joint position trajectories show the feedback reception times relative to the command publication times, with the step command applied at $t=0$.}
    \label{fig:latency}
\end{figure}

To address the sim-to-real transfer challenges in humanoid robots, we systematically quantify the dynamics gap between simulation and physical deployment, while adjusting our training setup accordingly. Fig. \ref{fig:sim2real} illustrates joint position-velocity trajectories during 10-second walking with a 0.5 m/s forward velocity command. By comparing results from training in Isaac Gym and the deployment in real world, we demonstrate that our carefully designed domain randomization effectively covers real-world properties and bridges the sim-to-real gap, enabling zero-shot policy transfer to physical robots. Additionally, cross-simulation results from MuJoCo indicate that the MuJoCo environment serves as an effective testing platform that closely approximates real-world dynamics. Its lightweight, cross-platform nature allows for rapid policy validation while maintaining consistency with physical robot performance.

To enhance sim-to-real alignment, we quantitatively quantify critical real-world parameters for domain randomization. For instance, we measure communication latency by sending joint commands and reading joint position feedback via Python SDK on the robot's onboard CPU. As shown in Fig. \ref{fig:latency}, our measurements indicate a round-trip communication latency of 9-12 ms from command transmission to the initial observed change in joint position. Additionally, we measure that the policy inference time is less than 1 ms. To better replicate these real-world temporal characteristics, we incorporate sensor-to-actuator latency randomization ranging from 0-20 ms during training in simulation. These quantified refinements ensure that our training scenarios comprehensively cover real-world dynamics. The synergistic implementation of these technical enhancements facilitates successful zero-shot policy transfer.

\section{Conclusions}

In this work, we introduce Booster Gym, an end-to-end reinforcement learning framework for humanoid robot locomotion that provides a complete pipeline from training to deployment with zero-shot sim-to-real transfer capabilities. Through the integration of carefully designed domain randomization techniques and environment configurations, the framework enables robust policy generalization across diverse terrains and conditions, demonstrating its practicality for real-world deployment. As an open-source solution, Booster Gym aims to lower the barrier for developing customized locomotion strategies and to foster community-driven innovation in humanoid robotics.

\addtolength{\textheight}{-12cm}   




\section*{Acknowlegement}

The development of Booster Gym draws inspiration from several open-source repositories, including \texttt{\href{https://github.com/isaac-sim/IsaacGymEnvs}{IsaacGymEnvs}} \cite{makoviychuk2021isaac}, \texttt{\href{https://github.com/leggedrobotics/legged_gym}{legged\_gym}} \cite{rudin2022learning}, \texttt{\href{https://github.com/leggedrobotics/rsl_rl}{rsl\_rl}}, and \texttt{\href{https://github.com/roboterax/humanoid-gym}{humanoid-gym}} \cite{gu2024humanoid}. These resources were referenced throughout the implementation process to enhance the design of our framework.


\bibliographystyle{IEEEtran}
\bibliography{IEEEabrv,IEEEexample}

\end{document}